\documentclass{article}
\usepackage{spconf,amsmath,graphicx}
\usepackage{hyperref}

\usepackage{cite}
\usepackage{amsmath,amssymb,amsfonts}
\usepackage{algorithmic}
\usepackage{graphicx}
\usepackage{textcomp}
\usepackage{xcolor}
\usepackage{algorithm}

\usepackage{caption}
\usepackage{titlesec}

\titleformat{\section}[block]{\normalfont\Large\bfseries\filcenter}{\thesection}{1em}{}
\titlespacing*{\section}{0pt}{1pt}{1pt}
\titlespacing*{\subsection}{0pt}{1pt}{1pt}

\title{A Self-supervised Pressure Map human keypoint Detection Approch: Optimizing Generalization and Computational Efficiency Across Datasets}
\name{Chengzhang Yu$^{1 2}$ \qquad Xianjun Yang$^{2}$ \qquad Wenxia Bao$^{1}$ \qquad Shaonan Wang$^{2}$ \qquad Zhiming Yao$^{2 \star}$}
\address{$^{1}$ Anhui University \\ $^{2}$ Hefei Institutes of Physical Science, Chinese Academy of Sciences}

\begin{document}
\topmargin=0mm
%
\maketitle

\begin{center}
\textcolor{red}{\textit{"This work has been submitted to the IEEE for possible publication. Copyright may be transferred without notice, after which this version may no longer be accessible."}}
\end{center}

\begin{abstract}
In environments where RGB images are inadequate, pressure maps is a viable alternative, garnering scholarly attention. This study introduces a novel self-supervised pressure map keypoint detection (SPMKD) method, addressing the current gap in specialized designs for human keypoint extraction from pressure maps. Central to our contribution is the Encoder-Fuser-Decoder (EFD) model, which is a robust framework that integrates a lightweight encoder for precise human keypoint detection, a fuser for efficient gradient propagation, and a decoder that transforms human keypoints into reconstructed pressure maps. This structure is further enhanced by the Classification-to-Regression Weight Transfer (CRWT) method, which fine-tunes accuracy through initial classification task training. This innovation not only enhances human keypoint generalization without manual annotations but also showcases remarkable efficiency and generalization, evidenced by a reduction to only $5.96\%$ in FLOPs and $1.11\%$ in parameter count compared to the baseline methods. Code is accessible at \href{https://anonymous.4open.science/r/SPMKD-52CB/}{SPMKD-52CB}.
\end{abstract}

\section{Introduction}
\label{Introduction}

In the rapidly evolving field of human image analysis, keypoint detection has emerged as a cornerstone, notably enhancing the accuracy and swiftness of image interpretation. Particularly, pressure maps, generated from data harvested by pressure mattresses, have begun to gain increasing attention in this domain, offering a rich application landscape for human keypoint detection. These maps have notably propelled advancements in medical and health monitoring, finding pivotal roles in recumbency recognition, body movement recognition, and pinpointing sleep-related breathing disorders\cite{huangUsingPressureMap2014,liHandMotionRecognition2014,guerrero2013detection}. Not only do they excel in safeguarding privacy and minimizing interference, a growing necessity in sensitive sectors, which fill a gap where RGB images fall short\cite{orekondyVisualPrivacyAdvisor2017,liuPixelPrivacy2019,climent-perezProtectionVisualPrivacy2021,lee2015changes,black2007national}. 

However, in the realm of pressure map-based human keypoint detection, progress lagged behind that of RGB-based images, with no specific models developed until around 2020. Liu et al.\cite{liuSimultaneouslyCollectedMultimodalLying2022} marked a significant milestone not only by presenting a dataset of pressure maps accompanied by accurate 2D human keypoint labels but also by successfully adapting various human keypoint detection algorithms to the domain of pressure maps, including adaptations of SHGlass\cite{liuSimultaneouslyCollectedMultimodalLying2022}, PoseAttention\cite{chu2017multi}, PyraNet\cite{yang2017learning}, HRpose\cite{sun2019deep}, and RESpose\cite{xiao2018simple}. These models, though innovative, were not without their shortcomings. Due to the inherent differences between pressure maps and RGB images, the migrated models often encountered issues of ambiguity in 2D or 3D poses, particularly when body parts lost contact with pressure sensors, resulting in human keypoint loss\cite{clever20183d,clever2020bodies}. This scenario underscores the necessity to develop new human keypoint detection strategies that are more aligned with the specific characteristics of pressure maps. Moreover, the prevalent reliance on manual human keypoint annotations, a notably challenging task even for humans in the context of pressure maps, underscores the necessity for a more customized approach.

In this study, we introduce a novel self-supervised pressure map human keypoint detection (SPMKD) approach, a synergy of the Encoder-Fuser-Decoder (EFD) model and the Classification-to-Regression Weight Transfer (CRWT) method, aimed at optimizing both generalization and computational efficiency across various tasks and datasets. The EFD model adeptly extracts and transforms human keypoints and their features from pressure maps, followed by a reconstruction of the pressure map through a Decoder. This strategy employs a lightweight Encoder for real-world applications, paired with a more substantial Decoder during the training phase, maintaining a minimalistic yet effective design. 

Addressing the challenge of acquiring large-scale data, which is often hindered by privacy concerns and the necessity for specialized resources, we present the CRWT method. This innovative technique begins with training the Decoder to classify pixel points based on the presence of pressure values on the map, utilizing these classification weights as a pre-training foundation for the regression task, thereby facilitating model convergence. This approach circumvents the need for extensive data collection, enhancing the efficiency and effectiveness of the EFD training process.

Our proposed approach, SPMKD,outperforms five manually annotated human keypoint detection ones. Compared with manually annotated RGB image human keypoint detection approaches, our model achieves higher accuracy, requiring only $5.96\%$ and $1.11\%$ of the number of FLOPs and parameters, respectively, while providing better generalization performance on unfamiliar datasets. The demonstrated potential of this model suggests that it could unlock the full potential of pressure maps across various applications.

The main contribution of this paper lies in the innovative introduction of the SPMKD approach, encompassing both EFD and CRWT components, adeptly addressing the two prominent challenges that have been plaguing the field of human keypoint extraction based on pressure maps.

\section{Method}
\label{sec:Method}

The research introduces the SPMKD method, consisting of two main components: the Encoder-Fuser-Decoder (EFD) model (Section \ref{Encoder}-\ref{Decoder}), which facilitates end-to-end training by extracting human keypoints and features from pressure maps and optimizing network weights, and the Classification-to-Regression Weight Transfer (CRWT) method (Section \ref{Training mode}), which eases the training process.

\begin{figure}[htbp]
  \centering
  \includegraphics[width=0.5\textwidth]{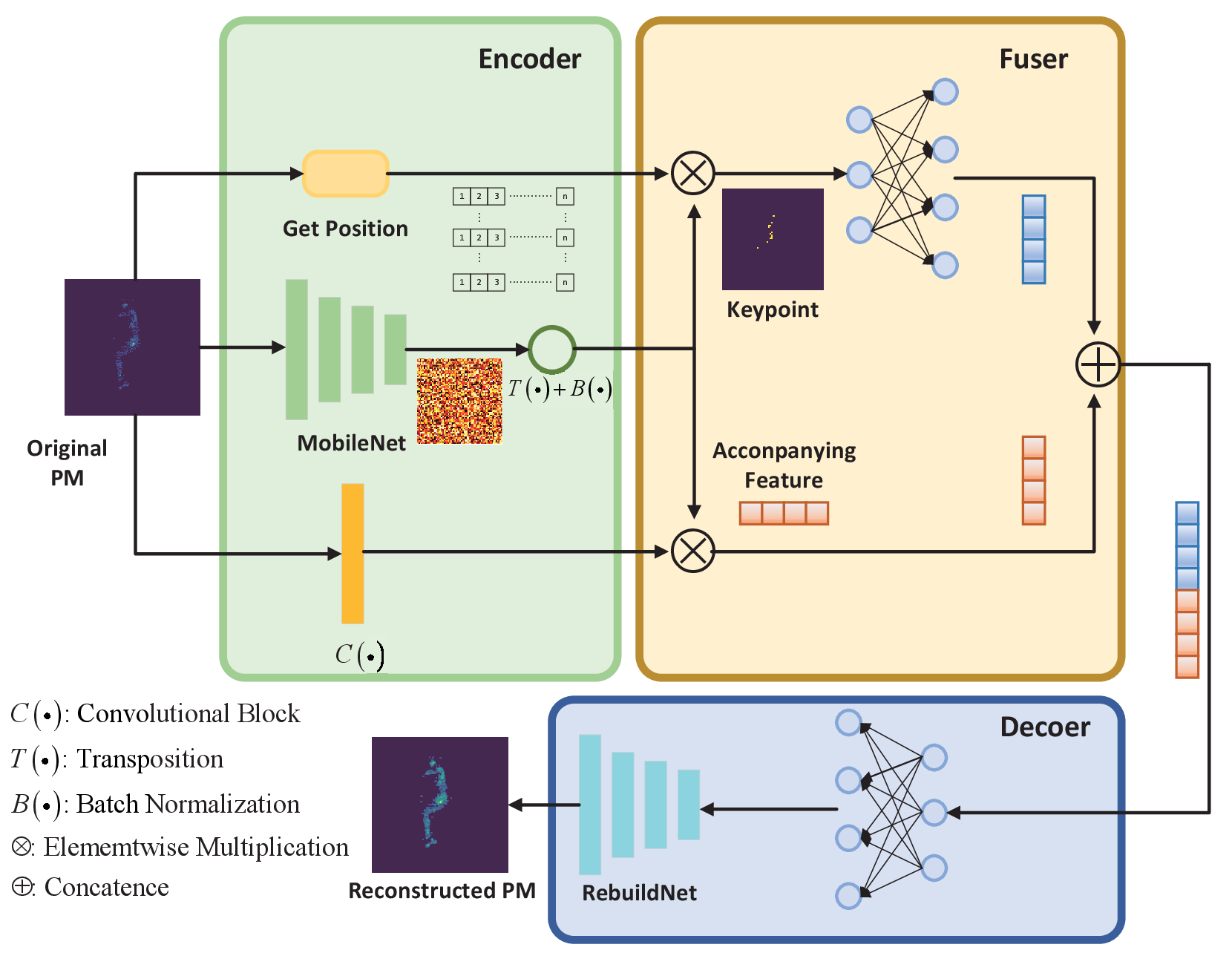}
  \caption{Overview of EFD: Encoder detects Keypoint Heatmaps, Positive Vectors, and Feature Vectors. Fuser fuses elements and applies dimensionality expansion to avoid weight shift. Decoder uses a fully connected layer and the proposed RebuildNet to reconstruct the pressure map.}
  \label{fig3}
\end{figure}

\subsection{Encoder}
\label{Encoder}

The Encoder is responsible for extracting human keypoints and their associated features, setting the stage for the classifier in the sub-task. For optimal performance, the Encoder should: 1) effectively discern human keypoints and features, and 2) maintain computational and parametric efficiency.

To meet these criteria, we adopt a dual-strategy. Initially, the input $X\in\mathbb{R}^{1\times256\times256}$ is transformed into a human keypoint heatmap $H\in\mathbb{R}^{h\times64\times64}$ using MobileNet V3\cite{howard2019searching}. Here, $h_{i,j,k}$ denotes a pixel of $H$, with $i$ indicating the probability of the point being the $i^{th}$ human keypoint.

Subsequently, a convolutional layer combined with basic mathematical operations extracts pixel features and locations, outputting as $F\in\mathbb{R}^{f\times64\times64}$ and $C\in\mathbb{R}^{3\times64\times64}$. $f$ represents the feature vector length corresponding to each pixel. By segregating heatmap, features, and locations, our model minimizes redundant parameters, ensuring efficient feature extraction vital for precise human keypoint detection in practical scenarios.

\subsection{Fuser}
\label{Fuser}

\begin{algorithm}
\caption{algorithm of Fuser}
\label{code1}
\begin{algorithmic}[1]
\REQUIRE \parbox[t]{0.8\linewidth}{$\textbf{P}$:Heat map for representing keypoints probability with shape $(k,w,h)$. \\
$\textbf{C}$:Position encoded vector with shape $(w,h,3)$ \\
$\textbf{F}$:Feature vector of pixel points with the shape $(w,h,f)$.}
\ENSURE Feature vector, called Output, with $(k, 2f)$ is used for subsequent reconstruction of the feature map.
\STATE $K_C = H *C$ \text{//} \textit{Matrix P and matrix $C$ implement matrix multiplication to obtain the location of keypoints.}
\STATE $K_F = H *F$ \text{//} \textit{Matrix P and matrix $F$ implement matrix multiplication to obtain the feature of keypoints.}
\STATE $K_C^{'} = f_l(K_C) $ \text{//} \textit{Convert $K_C$ to the same shape as $K_F$ by means of a fully connected layer.}
\STATE $Output = f_{cat}(K_c^{'},K_f)$  \text{//} \textit{Concatenate $K_C^{'}$ and $K_F$ together.}
\RETURN $Output$
\end{algorithmic}
\end{algorithm}

End-to-end training stands as a pivotal strategy in optimizing deep learning networks. However, the Encoder's output, a heatmap illustrating the probability of each pixel being a human keypoint, necessitates the extraction of human keypoint locations and features to facilitate input for the Decoder. Directly utilizing the highest value on the heatmap to pinpoint human keypoint locations can disrupt the gradient flow. To circumvent this, we introduce a method that calculates weighted locations instead of absolute value locations, preventing gradient interruption. This involves applying a sigmoid function to the heatmap $H$ to ascertain the relative importance of each pixel point $p_{h\in H}$. Consequently, the weighted heatmap aids in determining the weighted positions $C={c_{h\in H}\in\mathbb{R}^2}$ and features $F={f_{h\in H}\in\mathbb{R}^f}$ for each pixel, facilitating the identification of weighted locations and corresponding features. The specific process is shown in algorithm \ref{code1}.

\subsection{Decoder}
\label{Decoder}

The primary role of the Decoder is to transform the human keypoints and accompanying features back into a pressure map. Given the sparsity of human keypoint features relative to the entire pressure map, we first apply a fully connected layer to convert the input into a $32\times32$ feature map. 

\begin{figure}[htbp]
  \centering
  \includegraphics[width=0.5\textwidth]{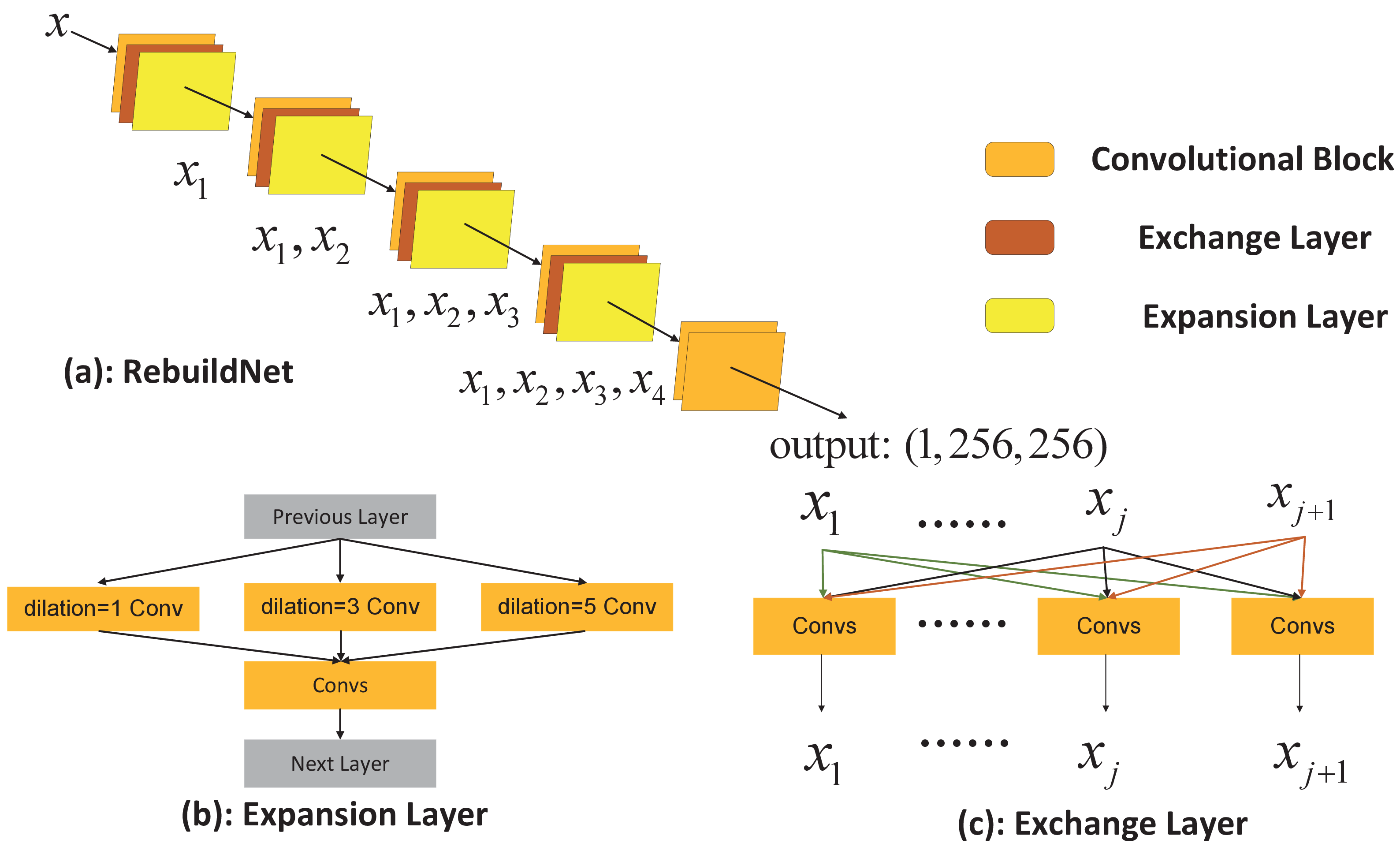}
  \caption{Structure of RebuildNet: (a) Overall structure, where the first four repetitive layers form the network backbone and the final two layers form the network head, interchangeable for different tasks. (b) Expansion Layer structure using three different sizes of dilated convolution for full fusion of features at different distances. (c) Exchange Layer structure ensures full fusion of features of different dimensions.}
  \label{fig4}
\end{figure}

Then, the Decoder, as illustrated in Fig. \ref{fig4}, functions through two main layers: the Expansion Layer and the Exchange Layer, progressively reconstructing the pressure map from a size of $64\times64$ to $256\times256$.

The Expansion Layer, equipped with several feature extraction and upsampling layers, employs three parallel dilated convolutional layers\cite{yu2015multi} with dilated rates of $1$, $3$, and $5$ to enhance upsampling accuracy amidst sparse input features. This layer focuses on refining features from the preceding layer to closely resemble the target pressure map.

Conversely, the Exchange Layer aims to mitigate gradient irregularities and reconstruction bias induced by network depth. It utilizes four parallel routes with widths of $64$, $128$, $256$, and $512$ elements for feature extraction, incorporating features from all paths at each stage to maintain a balance between different resolution features.

By adhering to these designs, the Decoder adeptly reconstructs the pressure map from sparse human keypoints and features, fostering precise and reliable human keypoint detection, thereby advancing pressure map analysis.

\subsection{Classification-to-Regression Weight Transfer (CRWT)}
\label{Training mode}

\begin{figure}[htbp]
\centering
\includegraphics[width=0.5\textwidth]{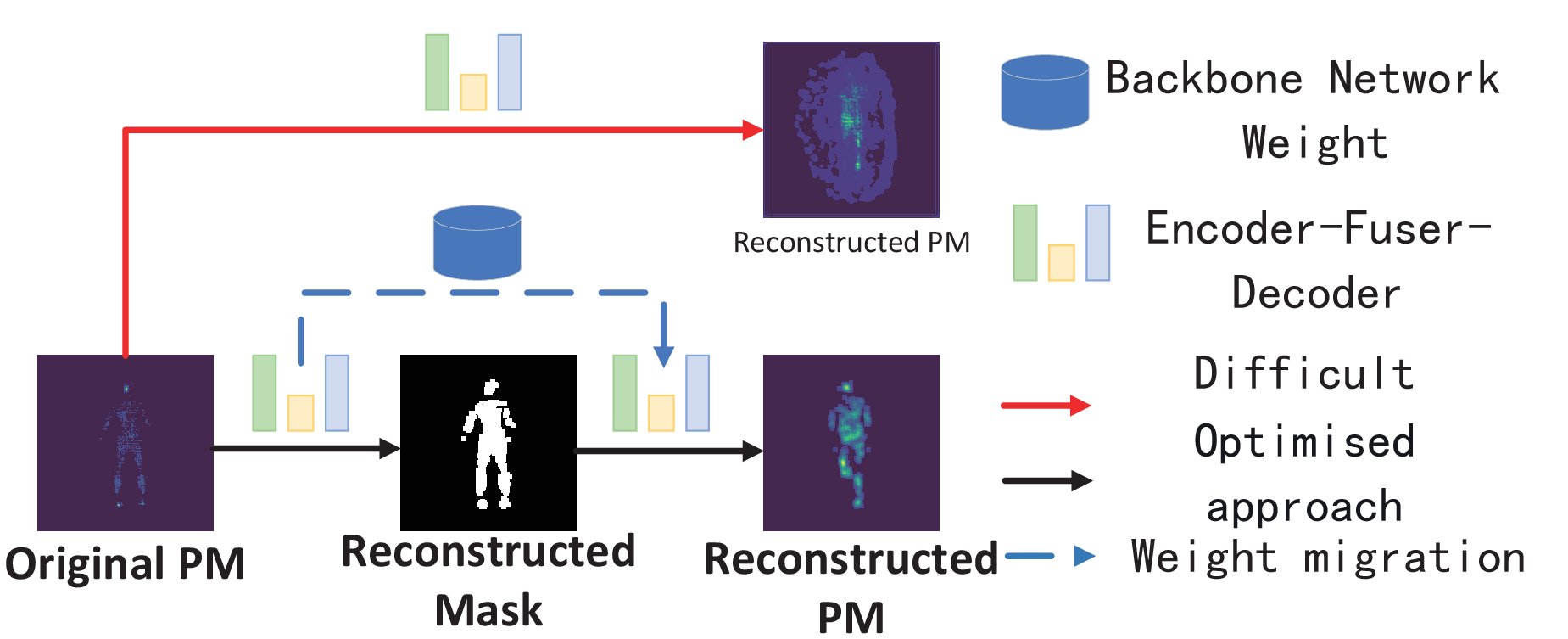}
\caption{CRWT Schematic: The red line illustrates the negative impact of omitting CRWT in pre-training weights, leading to significant noise in the training outcomes. In contrast, the black line represents the application of CRWT, demonstrating its effectiveness in improving training results.}
\label{fig6}
\end{figure}

To effectively address the challenges in acquiring large-scale datasets and to diminish the necessity for extensive data for neural network pre-training, we introduce the Cascaded Re-Weighting Training (CRWT) method. This novel approach leverages the weights from initial training as a basis for subsequent phases, thereby reducing the dependency on large datasets and enhancing the Encoder-Fuser-Decoder model's performance. The CRWT method's workflow is comprehensively depicted in Fig. \ref{fig6}.

Initially, the network is configured to predict the probability of pressure presence at each pixel point, simplifying the task from a complex regression to a more straightforward classification problem. This initial phase modifies the final Decoder layer's output to a dimension of $(2,64,64)$, with one channel indicating the likelihood of pressure presence and the other indicating its absence. This modification results in an improved f-score of $93.51\%$.

Subsequently, in the next phase of CRWT, the output dimension is reverted to $(1,64,64)$. The pre-training weights from the initial phase are then applied to enhance the model's reconstruction accuracy. This application significantly reduces the L2 and SSIM loss metrics, defined as follows:

\begin{align}
L &= \alpha L_2+\beta L_{S S I M} = \alpha L_2+\beta (1-SSIM) \\
SSIM &= 1-\frac{\left(2 \mu_H \mu_{H^{\prime}}+c_1\right)\left(2 \sigma_{H H^{\prime}}+c_2\right)}{\left(\mu_H^2+\mu_{H^{\prime}}^2+c_1\right)\left(\sigma_H^2+\sigma_{H^{\prime}}^2+c_2\right)}
\end{align}

In this formulation, $\alpha$ and $\beta$ represent the assigned weights to L2 loss and SSIM loss, respectively. The variables $p$ and $\mu$ denote the pixel values and mean values of the pressure maps. The term $c$ is a small positive constant introduced to stabilize the division process, and $n$ signifies the total number of pixel points.

Through these stages, CRWT method effectively bridges the gap between classification and regression tasks in neural network training, thereby significantly enhancing the overall performance and accuracy of the Encoder-Fuser-Decoder model.

\section{Experiment}
\label{sec:Experiment}


The results, detailed in Tab. \ref{table2}, underscore the superior accuracy of our proposed method compared to manual annotation-based techniques. Notably, this was achieved with a mere $5.96\%$ and $1.11\%$ of the FLOPs and parameters, respectively, amplifying its potential for practical application and adaptability in real-world scenarios.

Delving deeper, we conducted ablation experiments to scrutinize the performances of both manually annotated and self-supervised human keypoints in unfamiliar datasets, a necessary step given the distinct challenges pressure maps pose when compared to RGB images. Our analysis, outlined in Tab.\ref{table4}, revealed a significant dip in accuracy across all manually annotated models when transitioning to new datasets, with the RESpose\cite{xiao2018simple} model witnessing a substantial $26.77\%$ decline. In stark contrast, our self-supervised human keypoint approach exhibited remarkable adaptability, bolstering the accuracy of GAT\cite{kipf2016semi} and GSAGE\cite{hamilton2017inductive} classifiers by $3.21\%$ and $3.66\%$ respectively, thereby hinting at a promising avenue for enhanced generalization in unfamiliar datasets.

\begin{figure}[htbp]
  \centering
  \includegraphics[width=0.5\textwidth]{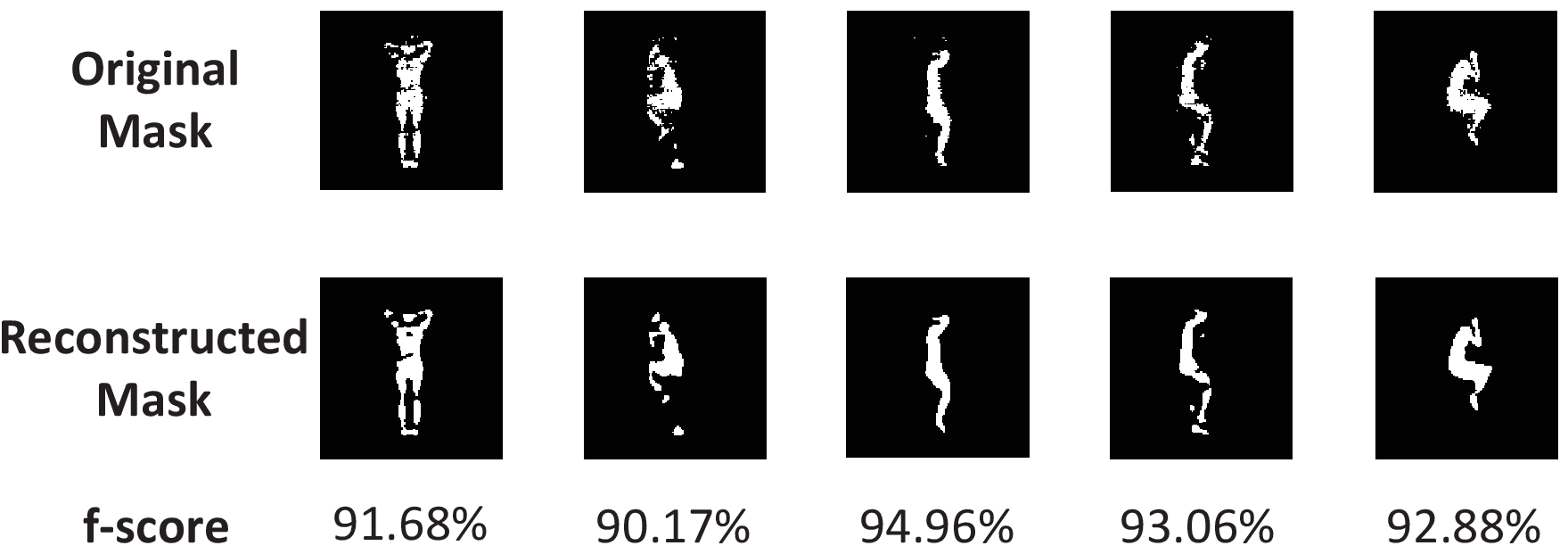}
  \caption{Pressure Map vs. Reconstructed Pressure Map: White areas are points with pressure values, and black areas are points without pressure values.}
  \label{fig5}
\end{figure}

\begin{figure}[htbp]
  \centering
  \includegraphics[width=0.5\textwidth]{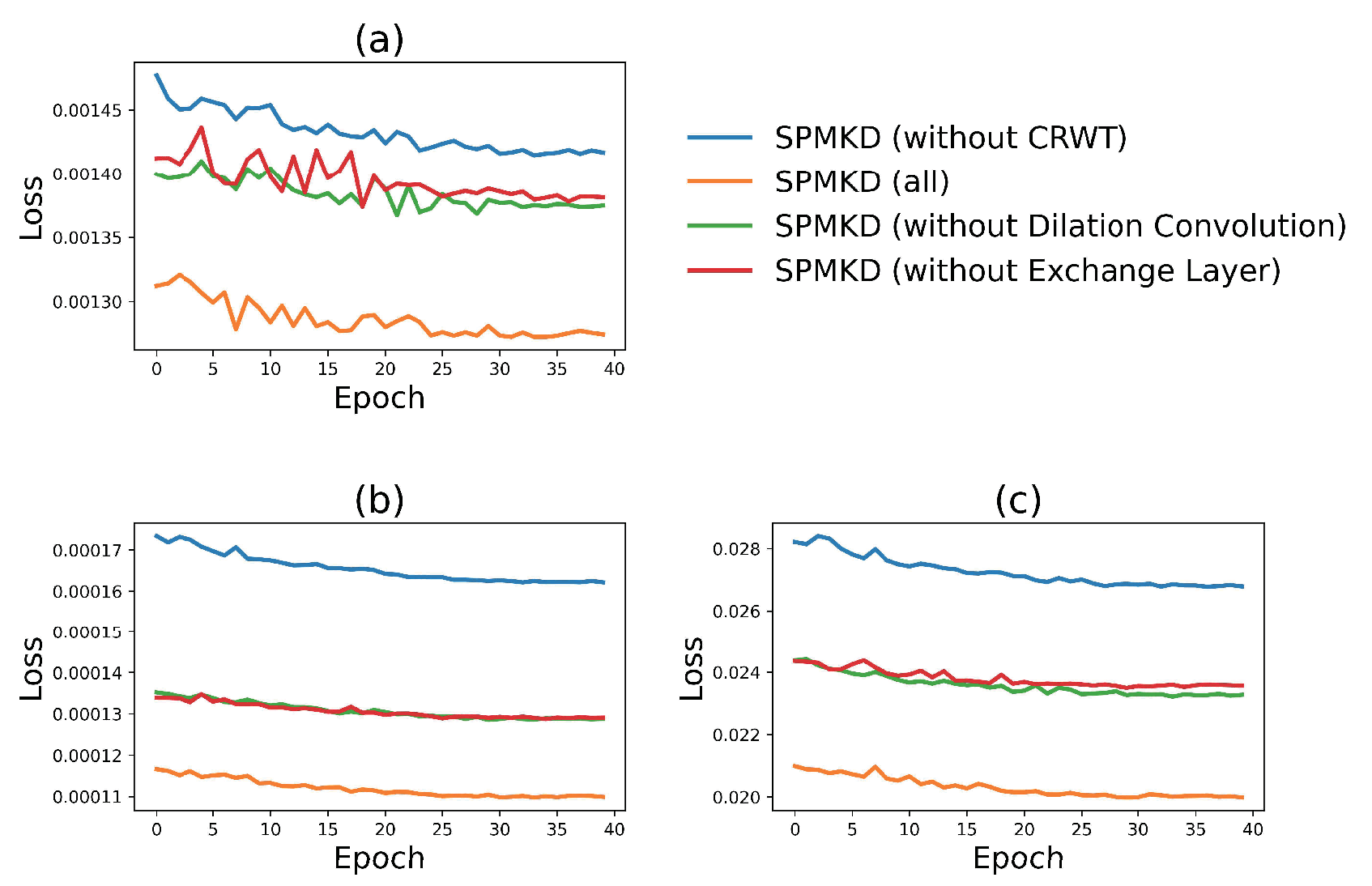}
  \caption{Loss Line Graph: For improved clarity, only the final 40 epochs are depicted. (a) L1 loss convergence curve. (b) L2 loss convergence curve, where SPMKD (excluding Dilation Convolution) partially overlaps with SPMKD (excluding Exchange Layer). (c) SSIM loss convergence curve.}
  \label{fig9}
\end{figure}

Further, we introduced the CRWT method, a novel initiative aimed at generating pre-training weights to mitigate training challenges. Our ablation study, utilizing two distinct loss functions as evaluative criteria, demonstrated that the CRWT method significantly alleviated convergence difficulties and augmented the model's reconstruction capabilities, as evidenced in Fig. \ref{fig9} (blue and orange lines). Fig. \ref{fig5} vividly illustrates the positive impact of CRWT on pressure point classification.

\begin{figure}[htbp]
  \centering
  \includegraphics[width=0.5\textwidth]{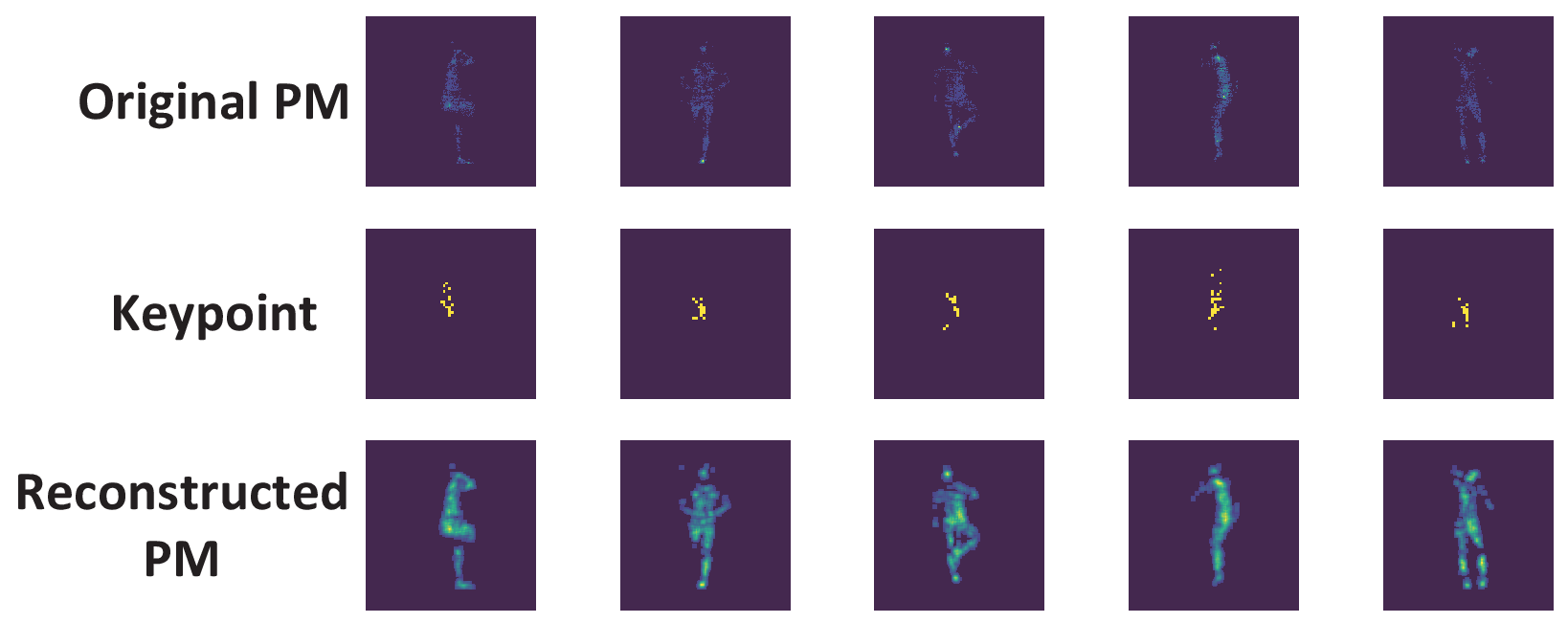}
  \caption{Pressure Map vs. Reconstructed Pressure Map: Lighter colors indicate greater pressure.}
  \label{fig2}
\end{figure}

Lastly, we explored the influence of the expansion and switching layers within the RebuildNet through an ablation analysis. Our results, highlighted in Fig. \ref{fig9} (orange, green, and red lines), affirm that the integration of these layers substantially enhances the reconstruction ability of the system. Fig. \ref{fig2} further showcases the impressive outcomes of the fully-integrated RebuildNet on the reconstructed pressure image results, cementing the potential of our approach in revolutionizing pressure map analysis.

\begin{table}
  \centering
  \setlength{\tabcolsep}{1pt} 
  \caption{Comparison of Accuracy (\%) Using Various Graph Convolutional Networks (GCNs) on the SLP Dataset.}
  \label{table2}
  \begin{tabular}{cccccc}
    \hline
    \textbf{Model} & \textbf{GCN} & \textbf{GAT} & \textbf{GSAGE} & \textbf{Flops} & \textbf{Params}\\
    \hline
    SHGlass\cite{liuSimultaneouslyCollectedMultimodalLying2022} & 94.57\% & 95.00\% & 95.12\% & 13.869G & 6.320M \\
    PoseAttention\cite{chu2017multi} & 94.26\% & 94.63\% & 94.88\% & 14.12G & 6.34M \\
    PyraNet\cite{yang2017learning} & 95.00\% & 95.01\% & 95.25\% & 24.59G & 13.61M \\
    HRpose\cite{sun2019deep} & 95.19\% & 94.94\% & 95.19\% & 10.27G & 14.26M \\
    RESpose\cite{xiao2018simple} & 93.89\% & 94.57\% & 94.57\% & 17.70G & 26.49M \\
    \textbf{SPMKD(Ours)} & \textbf{95.25\%} & \textbf{95.18\%} & \textbf{95.56\%} & \textbf{0.613G} & \textbf{0.07M} \\
    \hline
  \end{tabular}
\end{table}

\begin{table}
  \centering
  \caption{Comparison of Accuracy (\%) Using Various Graph Convolutional Networks (GCNs) on the SMaL Dataset}
  \label{table4}
  \begin{tabular}{cccc}
    \hline
    \textbf{Model} & \textbf{GCN\cite{kipf2016semi}} & \textbf{GAT\cite{velickovic2017graph}} & \textbf{GSAGE\cite{hamilton2017inductive}} \\
    \hline
    SHGlass\cite{liuSimultaneouslyCollectedMultimodalLying2022}          & 72.39\%                    & 80.39\%                    & 93.94\%                          \\
    PoseAttention\cite{chu2017multi}             & 82.67\%                    & 85.28\%                    & 94.33\%                          \\
    PyraNet\cite{yang2017learning}                   & 73.06\%                    & 77.39\%                    & 82.83\%                          \\
    HRpose\cite{sun2019deep}                    & 77.00\%                    & 81.89\%                    & 90.89\%                          \\
    RESpose\cite{xiao2018simple}                   & 64.56\%                    & 66.11\%                    & 72.06\%                          \\
    \textbf{SPMKD(Ours)}             & \textbf{89.72\%}                    & \textbf{98.39\%}                    & \textbf{99.22\%}                          \\
    \hline
  \end{tabular}
\end{table}

\bibliographystyle{IEEEbib}
\bibliography{zotero}

\end{document}